\documentclass[journal]{IEEEtran}
\usepackage[T1]{fontenc}
\usepackage{graphicx}
\usepackage{booktabs}
\usepackage{amsmath,amssymb}
\usepackage{array}
\usepackage{multirow}
\usepackage{url}
\usepackage{cite}
\usepackage{stfloats}
\usepackage{placeins}
\usepackage[svgnames]{xcolor}
\usepackage[hidelinks]{hyperref}
\usepackage{orcidlink}
\graphicspath{{Figs/}}

\newcommand{\sAP}[1]{\ensuremath{\mathrm{sAP}^{#1}}}

\begin{document}

\bstctlcite{IEEEexample:BSTcontrol}

\title{MiLSD: A Micro Line-Segment Detector for Resource-Constrained Devices}

\author{
Parsa Hassani Shariat Panahi$^\pi$~\orcidlink{0009-0005-2912-3754},
Amir Hossein Jalilvand$^\pi$~\orcidlink{0000-0002-7641-6606},
and M. Hassan Najafi$^+$~\orcidlink{0000-0002-4655-6229} \\
\thanks{$^\pi$School of Computer Engineering, Iran University of Science and Technology, Tehran, Iran\\
$^+$Electrical, Computer, and Systems Engineering Department, Case Western Reserve University, OH, USA}
}

\maketitle

\begin{abstract}
Line segment detection is a key building block in visual SLAM, 3D reconstruction, and industrial inspection. Recent deep learning methods have greatly improved accuracy, yet even the smallest models require several megabytes of memory, exceeding low-cost MCU capacity. This work investigates the maximum achievable accuracy under a sub-megabyte budget. We propose MiLSD, a detector tailored for MCU-level constraints, and systematically compare three output representations within a compact fully-convolutional backbone.

Our study shows that the proposed F-Clip center-with-length-and-angle formulation learns most effectively at small model sizes. We find that 8-bit quantization preserves full-precision performance, while 4-bit quantization causes significant degradation, particularly in angle regression, with quantization-aware training recovering only part of the loss. With a one-megabyte activation budget and inference enhancements including sub-pixel decoding, test-time augmentation, and a lightweight verifier, MiLSD improves \(\mathrm{sAP}^{10}\) on ShanghaiTech Wireframe from \(10.6\) (25k parameters, \(0.25\) MB) to \(24.1\) within \(1\) MB. Rather than competing with GPU-scale parsers, we map the accuracy--memory trade-off across representations, bit-widths, capacities, and post-processing strategies for embedded vision systems.
\end{abstract}

\section{Introduction}\label{sec:intro}
\IEEEPARstart{L}{ine} segments are a primitive structural feature in computer
vision: the straight edges of walls, doors, buildings, and machined
parts~\cite{vongioi2010lsd}. They support SLAM, structure-from-motion,
vanishing-point estimation, lane and power-line detection, and industrial
inspection. While much recent progress has targeted GPU- or cloud-based
platforms, this work focuses on detection under the tight memory and compute
constraints of real-time embedded hardware.

Classical detectors such as LSD~\cite{vongioi2010lsd} (Line Segment Detector)
and EDLines~\cite{akinlar2011edlines} run on a CPU but find \emph{all} edges,
while modern learned wireframe parsers~\cite{zhou2019lcnn,xue2020hawp,dai2021fclip,gu2022mlsd}
recover only salient segments but require GPU- or phone-class compute.
Table~\ref{tab:rw} situates representative methods across this spectrum.
Classical detectors grow line-support regions from local gradients and validate
them statistically. LSD groups pixels with consistent gradient orientation and
accepts a segment if its number of false alarms is below one. EDLines reaches
comparable quality faster by chaining edge pixels into clean chains. ELSED~\cite{suarez2021elsed}
targets embedded CPUs for high frame rates. Their shared weakness: accuracy
degrades under blur, low contrast, and clutter, and runtime is content-dependent.
At its evaluated \(640\times480\) resolution, ELSED also requires several
full-frame gradient and edge buffers (\(\sim\)1.5--2\,MB) that exceed even the
1\,MB SRAM of an STM32H7, and since its edge walk is global and data-dependent,
it cannot be tiled and admits no static worst-case memory bound, unlike a
fixed-cost CNN.
Classical detectors have also been mapped to FPGAs and ASICs for deterministic
latency, but these implementations accelerate hand-designed gradient logic, not
neural networks on general-purpose microcontrollers.

ShanghaiTech Wireframe~\cite{huang2018wireframe} reframed line detection as a
learning problem. L-CNN~\cite{zhou2019lcnn} proposed junctions and verified
candidate lines. AFM~\cite{xue2019afm} used attraction fields, while
HAWP~\cite{xue2020hawp} combined holistic fields with endpoint verification.
ULSD~\cite{li2021ulsd} generalized across pinhole, fisheye, and spherical cameras,
and LETR~\cite{xu2021letr} uses transformers for direct line detection. L-CNN, HAWP,
ULSD, and LETR achieve \(\mathrm{sAP}^{10}\approx63\)--\(70\) (AFM, an earlier
field-based method, scores \(\approx24\)), but assume workstation-class memory.
The unifying observation is that classical detectors are light but find every
edge, while learned parsers are accurate but GPU-bound; no learned method yet
occupies the MCU column of Table~\ref{tab:rw}.

Even the lightest learned detector, M-LSD-tiny~\cite{gu2022mlsd}, requires
at least 78\,MB of runtime memory, orders of magnitude beyond the SRAM
budget of typical microcontrollers. For context, the STM32F746 provides only
320\,KB of SRAM and 1\,MB of flash~\cite{stm32f746}. On such devices, the
bottleneck is peak activation memory, not parameter storage. Prior work differs
in how segments are encoded at the network output. Heatmap methods require a
separate linker. Tri-point and center-with-displacement designs (TP-LSD~\cite{huang2020tplsd},
M-LSD~\cite{gu2022mlsd}) target mobile inference. F-Clip~\cite{dai2021fclip}
compresses each segment to center, length, and angle; we encode that angle as a
double-angle \((\cos2\theta,\sin2\theta)\). On microcontrollers, activations dominate SRAM
usage~\cite{lin2021mcunetv2,banbury2021micronets}. MCUNet~\cite{lin2020mcunet}
uses hardware-aware search, and MCUNetV2~\cite{lin2021mcunetv2} adds patch-based
inference. Integer quantization with
straight-through gradients~\cite{jacob2018quant,bengio2013ste} underpins PTQ and
QAT. Int8 is generally safe; sub-8-bit demands care~\cite{rusci2019memory,novac2021sensors}.
To our knowledge, no prior work combines these threads for MCU-scale line
segment detection.

This work investigates the maximum achievable accuracy under a sub-megabyte
memory budget. We study three axes: (i)~output representation: heatmap,
center-with-displacement, and F-Clip-style center-with-length-and-angle;
(ii)~quantization: full-precision, 8-bit, and 4-bit; and (iii)~inference
enhancements: sub-pixel decoding, test-time augmentation, and a lightweight
verifier.

We propose MiLSD, a detector designed for MCU-scale memory. With a 1\,MB
activation budget, MiLSD achieves \(\mathrm{sAP}^{10}=24.1\) on ShanghaiTech
Wireframe, improving over a \(0.25\) MB baseline at \(10.6\). Our quantization
study reveals that 8-bit inference preserves full-precision performance, while
4-bit quantization causes significant degradation, particularly in angle
regression, where quantization-aware training recovers only part of the loss.
This sensitivity has not been reported in prior work.

The main contributions are:
\begin{itemize}
\item A comparison of three output representations under extreme memory
constraints, identifying F-Clip-style as the most effective at small model
sizes.
\item A quantization study revealing angle regression in the
\((\cos 2\theta, \sin 2\theta)\) space as the most sensitive component to
bit-width reduction.
\item MiLSD, operating within 1\,MB memory while achieving
\(\mathrm{sAP}^{10}=24.1\) on ShanghaiTech Wireframe.
\item An accuracy--resource frontier characterizing trade-offs among capacity,
quantization, and post-processing.
\end{itemize}

Targeting MCU-scale detection, where memory is the binding constraint, we
show that meaningful wireframe quality is achievable within 1\,MB despite
GPU-class parsers remaining out of reach. Our study maps the accuracy--resource
trade-offs across representations, quantization, and post-processing.


The rest of the paper is organized as follows. Section~\ref{sec:method}
details the proposed architecture, including the three output representations,
the compact backbone, and the quantization-aware training pipeline.
Section~\ref{sec:setup} describes the experimental protocol, dataset, and
training hyperparameters. Section~\ref{sec:results} presents our empirical
findings on representation selection, quantization sensitivity, capacity
scaling, and comparisons to prior art. Section~\ref{sec:milsd} introduces the
full MiLSD system on the STM32H7, incorporating capacity scaling, sub-pixel
decoding, test-time augmentation, and a learned verification head.
Section~\ref{sec:conclusion} summarizes our contributions and discusses
directions for future work.

\begin{table*}[tbp]
\centering
\caption{Representative line segment detectors across classical, learned, and
efficient regimes (core ideas and platforms per the respective papers). The \emph{On MCU?}
column highlights that prior learned methods target GPU, phone, or FPGA platforms.}
\label{tab:rw}
\setlength{\tabcolsep}{3pt}
\small
\begin{tabular}{@{}p{2.8cm}p{6.5cm}ccl c@{}}
\toprule
\textbf{Work} & \textbf{Core idea} & \textbf{Learned} & \textbf{Platform} & \textbf{Output} & \textbf{On MCU?}\\
\midrule
LSD~\cite{vongioi2010lsd} & Gradient region-growing + \emph{a contrario} validation & no & CPU & segments & no\\
EDLines~\cite{akinlar2011edlines} & Edge-drawing + line fitting + Helmholtz validation & no & CPU & segments & no\\
ELSED~\cite{suarez2021elsed} & Edge-drawing with segment continuation (embedded-oriented) & no & CPU/ARM & segments & no\\
Wireframe-CNN~\cite{huang2018wireframe} & Junction$+$line heatmaps, dataset that started the field & yes & GPU & heatmaps & no\\
L-CNN~\cite{zhou2019lcnn} & Junction proposals + line verification head & yes & GPU & junctions & no\\
AFM~\cite{xue2019afm} & Attraction-field map regressed then squeezed to lines & yes & GPU & field & no\\
HAWP~\cite{xue2020hawp,xue2023hawpv2} & Holistic attraction field + endpoint verification & yes & GPU & 4D field & no\\
TP-LSD~\cite{huang2020tplsd} & Tri-points (center $+$ two endpoints), single-stage & yes & GPU & tri-point & no\\
F-Clip~\cite{dai2021fclip} & Fully-convolutional center $+$ length $+$ angle & yes & GPU & center+geo & no\\
DeepLSD~\cite{pautrat2023deeplsd} & Learned line attraction field + classical refinement & yes & GPU & field & no\\
LETR~\cite{xu2021letr} & Transformer, line entities without edge maps & yes & GPU & endpoints & no\\
M-LSD / -tiny~\cite{gu2022mlsd} & Compact MobileNetV2, center $+$ displacement (mobile) & yes & Phone/GPU & center+disp & no\\
LSDNet~\cite{teplyakov2022lsdnet} & Lightweight CNN front-end + classical LSD back-end & yes & GPU/CPU & segments & no\\
EvLSD-IED~\cite{wang2024evlsd} & Event-based LSD via image-to-event distillation & yes & Event cam & endpoints & no\\
Predecessor (FPGA)~\cite{paper1} & Modified Canny + step-length linking, hardware accelerator & no & FPGA & segments & n/a\\
\textbf{This work} & Tiny FCN, F-Clip output, int8 quantized & yes & \textbf{MCU} & center+geo & \textbf{yes}\\
\bottomrule
\end{tabular}
\end{table*}

\section{Proposed MiLSD design}\label{sec:method}

This section describes the design of MiLSD, a line-segment detector optimized
for microcontroller-scale memory. We first present three output
representations for encoding segments on a fixed grid, then describe the
compact backbone shared across all variants, followed by the quantization
strategy that enables int8 deployment, and finally the training pipeline and
deployment flow.

\subsection{Output Representations}\label{sec:repr}

\begin{figure}[!t]
\centering
\includegraphics[trim={1.6cm 2.2cm 1.5cm 1cm},clip,width=\columnwidth]{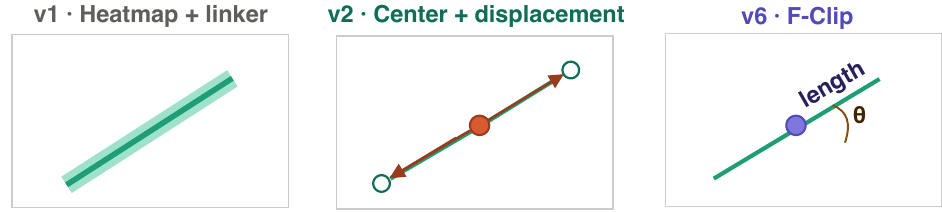}
\caption{Three output encodings as dense per-pixel maps (not explicit segments).
\textbf{v1}~predicts per-pixel line probability and requires an external linker.
\textbf{v2}~Center\,+\,displacement~\cite{huang2020tplsd,gu2022mlsd}: center
confidence with two endpoint offset vectors (four channels; int4-fragile).
\textbf{v6}~F-Clip~\cite{dai2021fclip}: center plus length and angle
\((\cos 2\theta, \sin 2\theta)\).}
\label{fig:representations}
\vspace{-1em}
\end{figure}

A key design decision for any learned line-segment detector is how to encode
the continuous geometry of a segment into discrete network outputs. We study
three representations on a \(128\times128\) output grid (Fig.~\ref{fig:representations}).

\emph{(i) Heatmap:} A per-pixel binary classification map indicating whether a
pixel lies on a line. This is the most direct representation but requires an
external post-processing linker to assemble pixels into continuous segments.
While conceptually simple, the linker introduces additional computational
overhead and hyperparameters, and the network itself does not produce
geometric primitives.

\emph{(ii) Center-with-Displacement:} Inspired by TP-LSD and M-LSD~\cite{huang2020tplsd,gu2022mlsd}, this
representation predicts a center confidence map alongside displacement vectors
from each center pixel to the two endpoints. Each segment is thus encoded as a
center point plus two offset vectors. This formulation is single-stage and
does not require an external linker, but the network must learn to regress
four continuous values (two displacements) per detected segment.

\emph{(iii) F-Clip:} Following Dai et al.~\cite{dai2021fclip}, this
representation encodes each segment as a center confidence map, a length
\(\ell\), and an orientation. Whereas F-Clip regresses the angle directly as a
scalar, we encode it as \((\cos 2\theta, \sin 2\theta)\); this double-angle
encoding resolves the \(180^\circ\) ambiguity inherent to undirected line
segments, making it uniquely defined for any line orientation.
For a ground-truth segment with endpoints \(\mathbf{p}_1,\mathbf{p}_2\), the
targets at the center cell are:
\[
\ell = \lVert\mathbf{p}_2 - \mathbf{p}_1\rVert, \quad
\theta = \operatorname{atan2}(y_2-y_1, x_2-x_1)
\]
During inference, decoding inverts this representation: for each detected
center peak above a confidence threshold, the segment is reconstructed as a
line of length \(\ell\) oriented at \(\theta\), centered at the detected
location. This compact encoding requires only four output channels (center,
length, \(\cos 2\theta\), \(\sin 2\theta\)), making it particularly attractive
for memory-constrained deployment. Section~\ref{sec:res-climb} compares these
three representations under identical memory budgets.

\subsection{Backbone Architecture}

\begin{figure}[!t]
\centering
\includegraphics[trim={0cm 0.5cm 0cm 0cm},clip,width=\columnwidth]{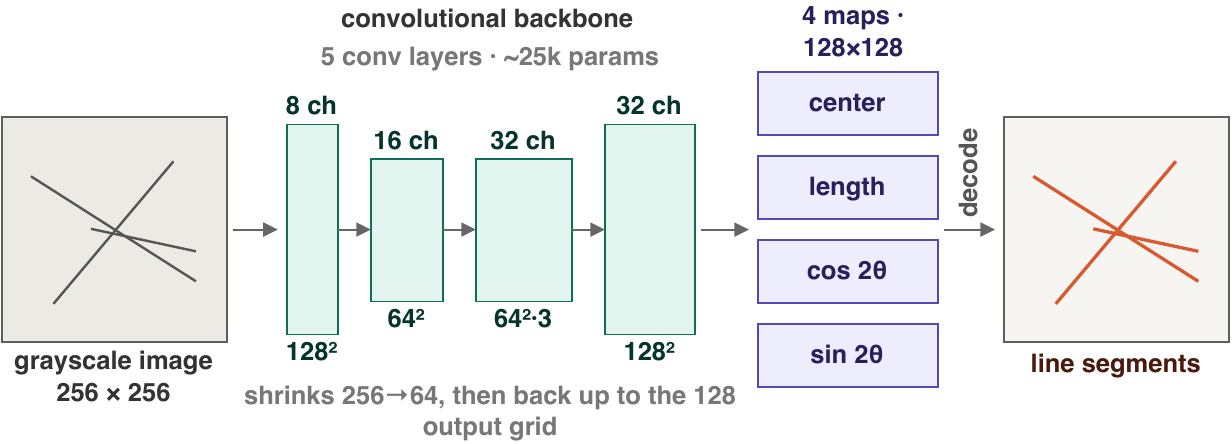}
\caption{Backbone and output head architecture. A \(256\times256\) grayscale
input is encoded through five strided convolutions, reducing spatial resolution
to \(64\times64\) and expanding channels to \(32\). A \(1\times1\) projection
and nearest-neighbor upsampling restore the resolution to \(128\times128\),
followed by a \(3\times3\) output head producing the four prediction maps.}
\label{fig:architecture}
\end{figure}

All three output heads share a common compact backbone designed to minimize
activation memory while preserving sufficient spatial resolution for accurate
line localization. The architecture (Fig.~\ref{fig:architecture}) consists of:

\begin{enumerate}
\item A strided fully-convolutional encoder with five convolutional layers,
channel widths \(8 \to 16 \to 32 \to 32 \to 32\), and stride-2 downsampling
that reduces the input resolution from \(256\times256\) to \(64\times64\).
\item A \(1\times1\) convolutional reduction layer that projects features to a
compact representation.
\item A nearest-neighbor upsampling layer that restores the spatial resolution
to \(128\times128\).
\item A \(3\times3\) output head that produces the final prediction maps
(center confidence, length, and orientation for F-Clip; center and
displacements for endpoint representation; or a single heatmap).
\end{enumerate}

The total parameter count is approximately \(25\)k at the smallest width.
Section~\ref{sec:res-cap} demonstrates that parameter capacity is not the
primary bottleneck in this regime; rather, input resolution and activation
memory constrain performance.

\subsection{Quantization for MCU Deployment}

\begin{figure}[!t]
\centering
\includegraphics[trim={2.5cm 0.5cm 1.5cm 0.5cm},clip,width=\columnwidth]{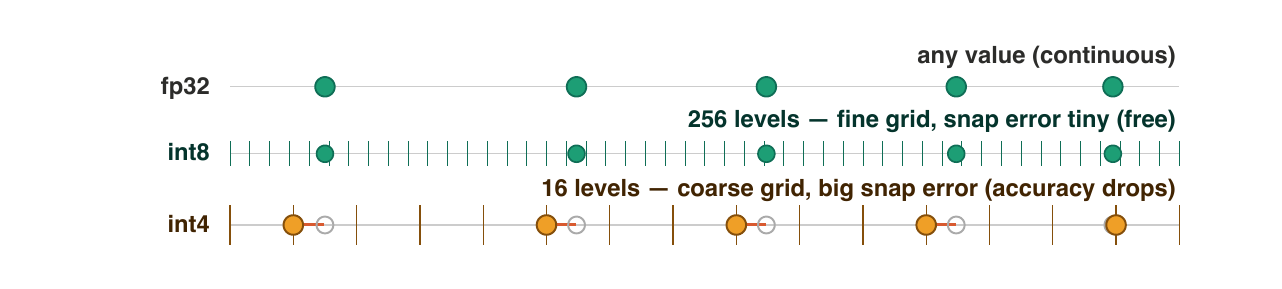}
\caption{Quantization scheme. Continuous weights are snapped to discrete levels:
int8 provides 256 levels (fine grid), while int4 provides only 16 levels
(coarse grid). Quantization is performed per-tensor and symmetric.}
\label{fig:quant_scheme}
\vspace{-1em}
\end{figure}

To fit the model within microcontroller SRAM and use optimized integer
inference kernels such as CMSIS-NN~\cite{lai2018cmsisnn}, we quantize weights
and activations to integer precision. We adopt per-tensor symmetric
quantization, implemented with fake-quantization in the forward pass and the
straight-through estimator (STE) for gradient propagation during
backpropagation~\cite{jacob2018quant,bengio2013ste}.

For a weight tensor \(w\) and target bit-width \(b\), the quantization scale is:
\[
s = \frac{\max|w|}{2^{b-1} - 1}
\]
The quantized weight is computed as:
\[
\hat{w} = s \cdot \operatorname{round}\left(\frac{w}{s}\right)
\]
This operation snaps continuous values onto a discrete grid of \(2^b\) levels.
As illustrated in Fig.~\ref{fig:quant_scheme}, 8-bit quantization provides 256
levels, offering fine granularity, while 4-bit quantization reduces this to
only 16 levels, introducing substantial rounding error.

We evaluate two quantization strategies:
\begin{itemize}
\item \emph{Post-training quantization (PTQ):} The model is first trained in
full precision, then weights and activations are quantized once using a small
calibration set. This is computationally efficient but can suffer from
accuracy degradation, particularly at low bit-widths.
\item \emph{Quantization-aware training (QAT):} The quantization operation is
simulated during training, allowing the model to learn weights that are robust
to quantization error. While more expensive, QAT often recovers some of the
accuracy lost in PTQ.
\end{itemize}

Our deployed model uses int8 quantization with QAT, achieving performance
comparable to full precision as shown in Section~\ref{sec:res-quant}. We also
investigate 4-bit quantization to understand the limits of aggressive
compression, revealing that angle regression is particularly sensitive to
bit-width reduction.

\subsection{Training Pipeline and Deployment Flow}

Training is performed entirely off-device on a GPU workstation; the
microcontroller executes only int8 inference. This train-off / infer-on split
is the defining premise of our deployment strategy and is common practice in
TinyML systems.

The training pipeline proceeds as follows:
\begin{enumerate}
\item A \(256\times256\) grayscale image is fed into the backbone.
\item The network produces prediction maps (center, length, \(\cos 2\theta\),
\(\sin 2\theta\) for F-Clip) on a \(128\times128\) grid.
\item Loss is computed against ground-truth segments encoded in the same
representation, using a combination of binary cross-entropy for center
confidence and smooth L1 loss for geometric attributes.
\item For QAT, quantization simulation is enabled during training with STE
gradient propagation.
\end{enumerate}

\begin{figure*}[!t]
\centering
\includegraphics[width=0.8\textwidth]{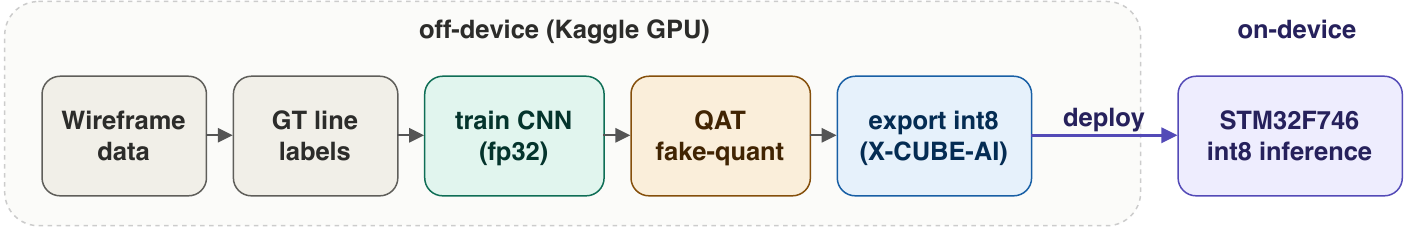}
\caption{Off-device training and on-MCU inference pipeline. The model is trained
on GPU with quantization simulation, then exported through X-CUBE-AI for
deployment on the STM32F746. The MCU executes int8 inference only.}
\label{fig:pipeline}

\end{figure*}

For deployment, the trained model is exported through X-CUBE-AI, STMicroelectronics'
neural network inference library for STM32 microcontrollers. The export process
generates optimized C code that runs on the Arm Cortex-M7 core, using
CMSIS-NN for efficient integer arithmetic. The inference pipeline on the MCU
(Fig.~\ref{fig:pipeline}) consists of:
\begin{enumerate}
\item Input image capture (grayscale, \(256\times256\)).
\item int8 inference through the quantized network.
\item Decoding of output maps into line segments (center detection,
length and angle extraction, endpoint computation).
\item Optional post-processing: Line-of-Interest verification and non-maximum
suppression.
\end{enumerate}

The entire inference pipeline is designed to operate within the 320\,KB SRAM
budget of the STM32F746, with peak activation memory as the primary constraint
rather than parameter storage.

\section{Experimental Setup}\label{sec:setup}
\paragraph{Dataset and metric}
We train and evaluate on the ShanghaiTech Wireframe
benchmark~\cite{huang2018wireframe} (5,000 training and 462 held-out evaluation
images, $\sim$74 segments per image). Accuracy is structural average precision $\sAP{t}$
at squared-endpoint-distance thresholds $t\in\{5,10,15\}$ in the $128^2$ output
space; we headline $\sAP{10}$. For comparability with the FPGA
predecessor~\cite{paper1} we also reference its Q1 (coverage) and Q2
(noise-suppression) measures.

\paragraph{Implementation}
Table~\ref{tab:hparams} lists the training configuration. Training is in PyTorch
on a GPU; versions v1--v6 of the design search~\cite{lsdtmlkaggle} share the
backbone of Section~\ref{sec:method}. All models are trained for 300 epochs with
a batch size of 32, using the Adam optimizer and a cosine annealing learning
rate schedule starting from $10^{-3}$. Data augmentation is limited to random
horizontal and vertical flips. The loss function combines binary cross-entropy
for center classification with smooth L1 loss for length and angle regression,
weighted by a factor of 2.0 for the geometric terms and masked to ground-truth
segment locations. We evaluate both PTQ and QAT at 8 and 4 bits.

\begin{table}[!t]
\caption{Training hyperparameters (deployed F-Clip model).}
\label{tab:hparams}
\centering
\small
\begin{tabular}{p{0.46\columnwidth} p{0.42\columnwidth}}
\toprule
Setting & Value\\
\midrule
Input $\to$ output resolution & $256\times256 \to 128\times128$\\
Backbone channels & $8\!\to\!16\!\to\!32\!\to\!32\!\to\!32$ ($\sim$25k params)\\
Output head & center, length, $\cos2\theta$, $\sin2\theta$\\
Epochs / batch & 300 / 32\\
Optimizer / LR & Adam / $1\times10^{-3}$, cosine schedule\\
Augmentation & horizontal $+$ vertical flip\\
Loss & BCE(center) $+\,2.0\times$ L1(length, angle), masked\\
Quantization & per-tensor symmetric; PTQ and QAT\\
\bottomrule
\end{tabular}
\end{table}
\section{Results}\label{sec:results}

\subsection{Representation Comparison: The Climb}\label{sec:res-climb}

\begin{figure}[!t]
\centering
\includegraphics[width=\columnwidth]{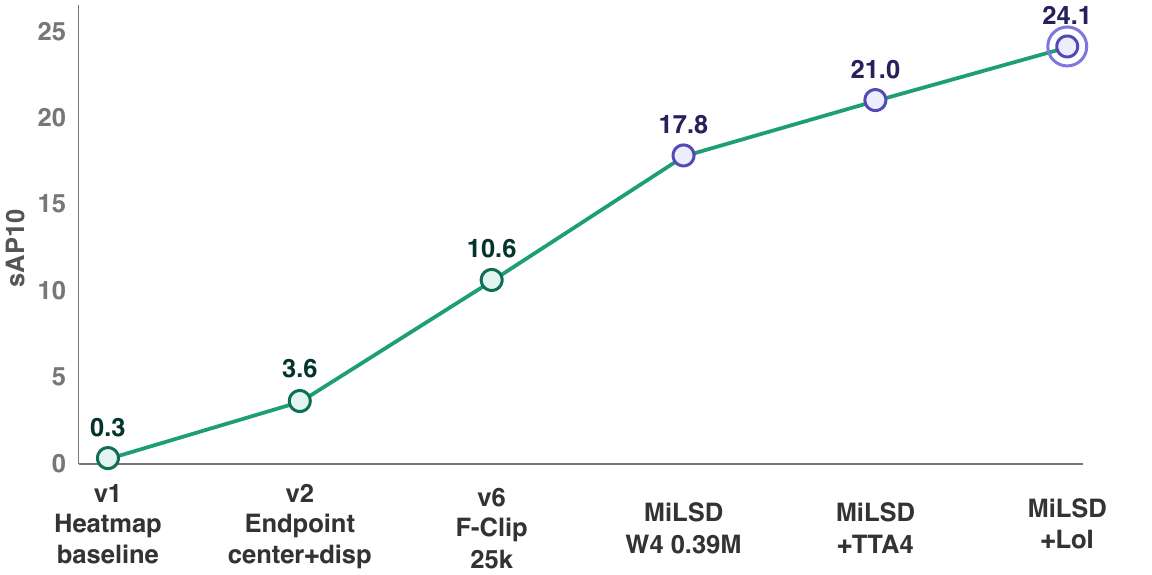}
\caption{From baseline to MiLSD: $\sAP{10}$ on Wireframe at each step. The output
representation drives the first gains (heatmap v1 $\to$ endpoint v2 $\to$ F-Clip),
reaching $10.6$ for the 25k-parameter F-Clip model; scaling capacity to the width-4
MiLSD backbone ($0.39$\,M parameters) lifts this to $17.8$, and the inference-time
stages (test-time augmentation then the Line-of-Interest verification head)
carry it to $\sAP{10}=24.1$ (purple).}
\label{fig:climb}
\end{figure}

Fig.~\ref{fig:climb} traces the evolution of $\sAP{10}$ across the successive
versions v1--v6 of our design search, providing a step-by-step account of how
accuracy accumulates as each design choice is introduced. At the lowest rung of
this progression, the heatmap baseline proves fundamentally inadequate for
producing clean, discrete segments, registering only $\sAP{10}=0.3$. Endpoint
regression constitutes the first formulation capable of yielding a meaningful
structural score, reaching $\sAP{10}=3.6$, yet it remains limited in its
ability to recover coherent geometry. The decisive inflection occurs with the
adoption of the F-Clip representation, which at the identical 25k-parameter
budget more than doubles performance to $\sAP{10}=7.2$, establishing that the
output encoding itself, rather than model size, is the dominant factor at this
scale. Subsequent refinements complete the climb in a more incremental fashion:
aligning the output grid with the label resolution at $256$~px input lifts
accuracy to $8.6$; incorporating the full training set of 5,000 images together
with flip augmentation raises it further to $9.3$; and extending the training
schedule to 300 epochs yields the final deployed score of $\sAP{10}=10.6$. Taken
together, these results demonstrate unambiguously that the largest gains
originate from the choice of \emph{representation}, not from additional
capacity. This finding carries particular significance for microcontroller-scale
design: a geometric encoding that explicitly parameterizes each segment by its
center, length, and orientation equips even a 25k-parameter network with
sufficient inductive structure to learn meaningful segment hypotheses, whereas
the heatmap and endpoint alternatives remain unable to assemble coherent
geometric predictions under the same severe parameter constraint.

\subsection{Quantization}\label{sec:res-quant}

\begin{figure}[!t]
\centering
\includegraphics[trim={2.3cm 1.2cm 2.5cm 1.2cm},clip,width=\columnwidth]{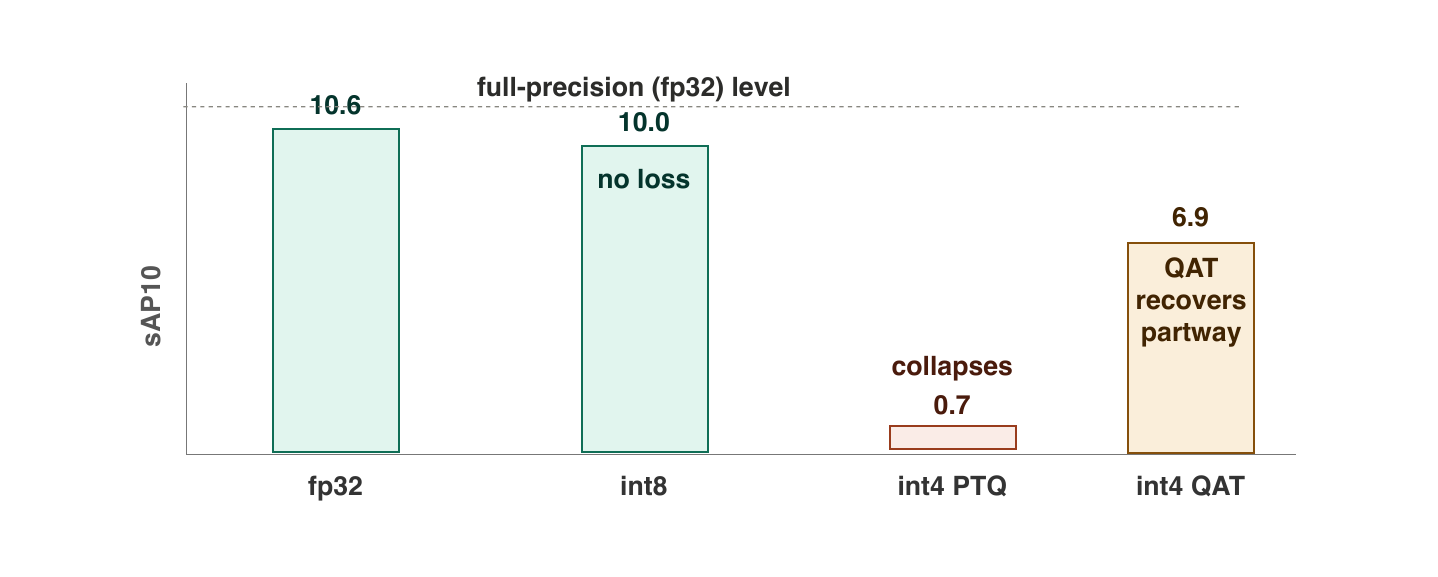}
\caption{$\sAP{10}$ versus bit-width for the deployed F-Clip model on Wireframe.
The int8 point overlaps fp32; PTQ at 4~bits fails while QAT partially recovers.}
\label{fig:quant_curve}
\end{figure}

Fig.~\ref{fig:quant_curve} and Table~\ref{tab:quant} present the results of our
systematic quantization study, evaluated across the three principal output
representations corresponding to versions v1, v2, and v6 of the design search.
For the deployed F-Clip model, the transition from full-precision fp32 inference
to 8-bit integer quantization incurs a degradation of only $0.6$
$\sAP{10}$ points ($10.6\rightarrow10.0$), indicating that int8 arithmetic is
sufficient for on-device deployment with negligible impact on structural
accuracy. By contrast, post-training quantization at 4 bits proves catastrophic,
collapsing performance to $\sAP{10}=0.7$; quantization-aware training partially
mitigates this failure, recovering the score to $6.9$, which corresponds to
approximately $60\%$ of the int4-induced gap relative to the fp32 baseline.
Inspection of the per-head errors reveals that the degradation is concentrated
almost entirely in the $(\cos2\theta,\sin2\theta)$ angle regression branch, whose
inherently narrow dynamic range is poorly served by the coarse 4-bit
quantization grid. On this basis, the deployed model adopts int8 throughout.
More broadly, these results suggest that pushing quantization beyond 8 bits is
unlikely to remain viable for geometric regression heads of this kind without
substantial architectural or training modifications.

\begin{table}[!t]
\caption{Quantization results, \sAP{10} on Wireframe.}
\label{tab:quant}
\centering
\small
\begin{tabular}{lcccc}
\toprule
Representation & fp32 & int8 & int4 PTQ & int4 QAT\\
\midrule
Heatmap (v1)$^{*}$ & 0.3 & 0.3 & 0.3 & n/a\\
Endpoint (v2) & 3.6 & 3.6 & $\sim$0.3 & fragile\\
\textbf{F-Clip (v6, deployed)} & \textbf{10.6} & \textbf{10.0} & 0.7 & 6.9\\
\bottomrule
\end{tabular}\\[2pt]
{\footnotesize $^{*}$Heatmap is not built for sAP; on its own terms Q2$=0.86$,
recall$=0.44$.}
\end{table}

\subsection{Resolution--Memory Trade-off}\label{sec:res-res}

\begin{figure}[!t]
\centering
\includegraphics[width=\columnwidth]{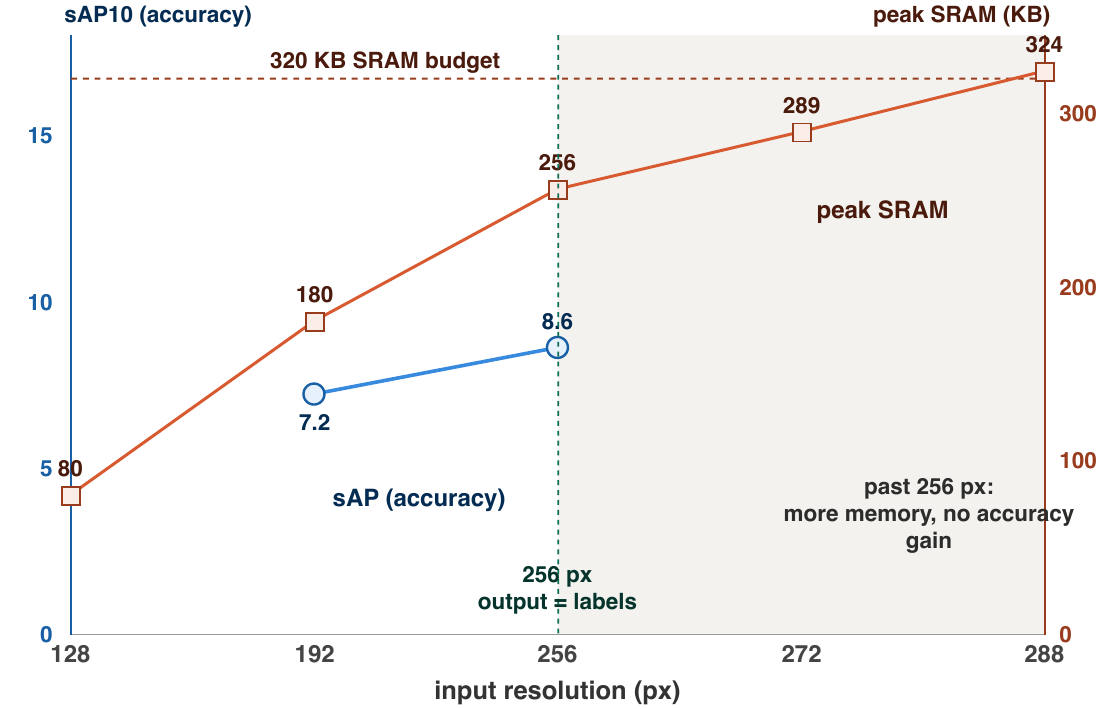}
\caption{Input resolution sets a genuine accuracy--memory trade-off. Accuracy (blue,
left axis) climbs as resolution rises, but only until the $128$ output grid reaches
the $128$-px label resolution at $256$\,px input; past that the output is finer than
the labels and accuracy saturates, while peak SRAM (coral, right axis) keeps growing
and crosses the $320$\,KB budget. $256$\,px is therefore the operating point.}
\label{fig:sram_resolution}
\end{figure}

Fig.~\ref{fig:sram_resolution} plots both $\sAP{10}$ and peak SRAM consumption as
functions of input resolution, enabling a joint assessment of the
accuracy--memory trade-off that governs operating-point selection on
resource-constrained hardware. The chosen configuration of $256$~px input is
selected at the point where the $128\times128$ output grid aligns with the
native label resolution, and where peak SRAM remains within the 320~KB ceiling
imposed by the STM32F746. This analysis reveals a pronounced and interpretable
trade-off: as input resolution increases, structural accuracy improves steadily
until the output grid reaches parity with the label resolution, at which point
further resolution gains yield diminishing or negligible returns. Beyond this
saturation point, peak SRAM continues to grow without a commensurate accuracy
benefit. The $256$~px operating point therefore represents the optimal balance
between detection quality and memory footprint for the F746 deployment target.

\subsection{Capacity and Overfitting}\label{sec:res-cap}

\begin{figure}[!t]
\centering
\includegraphics[width=0.9\columnwidth]{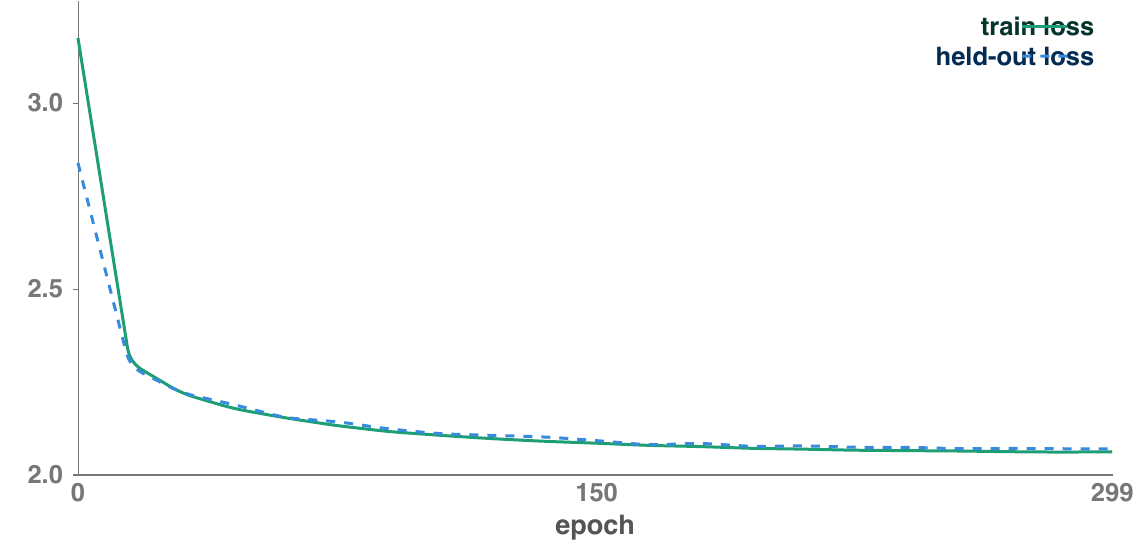}
\caption{Training the F-Clip model over 300 epochs. Train and held-out loss track
together throughout, with a final gap of $\approx0.01$: the 25k-parameter model
shows no overfitting despite only $5{,}000$ training images.}
\label{fig:training}
\end{figure}

To test whether the 25k-parameter backbone is capacity-limited, we swept
backbone width through three operating points ($\sim$25k, $\sim$98k, and
$\sim$209k parameters) while holding all other training settings fixed, and
recorded the held-out loss floor at convergence. The result is nearly
flat: the smallest model settles at $1.74$, while the intermediate and largest
variants both reach $1.73$; an eightfold increase in parameters yields only a
$0.01$ reduction in held-out loss. This pattern is consistent with the
capacity-gap effect documented in the knowledge-distillation
literature~\cite{mirzadeh2020takd}, and implies limited headroom for naive
distillation-based improvement at this scale. The flatness further suggests
that the model is already operating near the information-theoretic limit imposed
by the dataset and the chosen input resolution, such that additional parameters
are unlikely to translate into measurable gains in structural accuracy.
Complementing this capacity analysis, Fig.~\ref{fig:training} plots the training
and held-out loss trajectories over the full 300-epoch schedule. The two curves
remain closely aligned throughout training, converging to a final gap of
approximately $0.01$, indicating the absence of overfitting despite the severely
constrained 25k-parameter architecture and the modest size of the training set.
Together, these observations support the conclusion that the fixed small-width
design is both memory-efficient and well-regularized by its
architectural constraints.

\vspace{-1.25em}
\subsection{Comparison with Prior Work}\label{sec:res-cmp}

\begin{figure}[!t]
\centering
\includegraphics[trim={1cm 0.9cm 1cm 0cm},clip,width=\columnwidth]{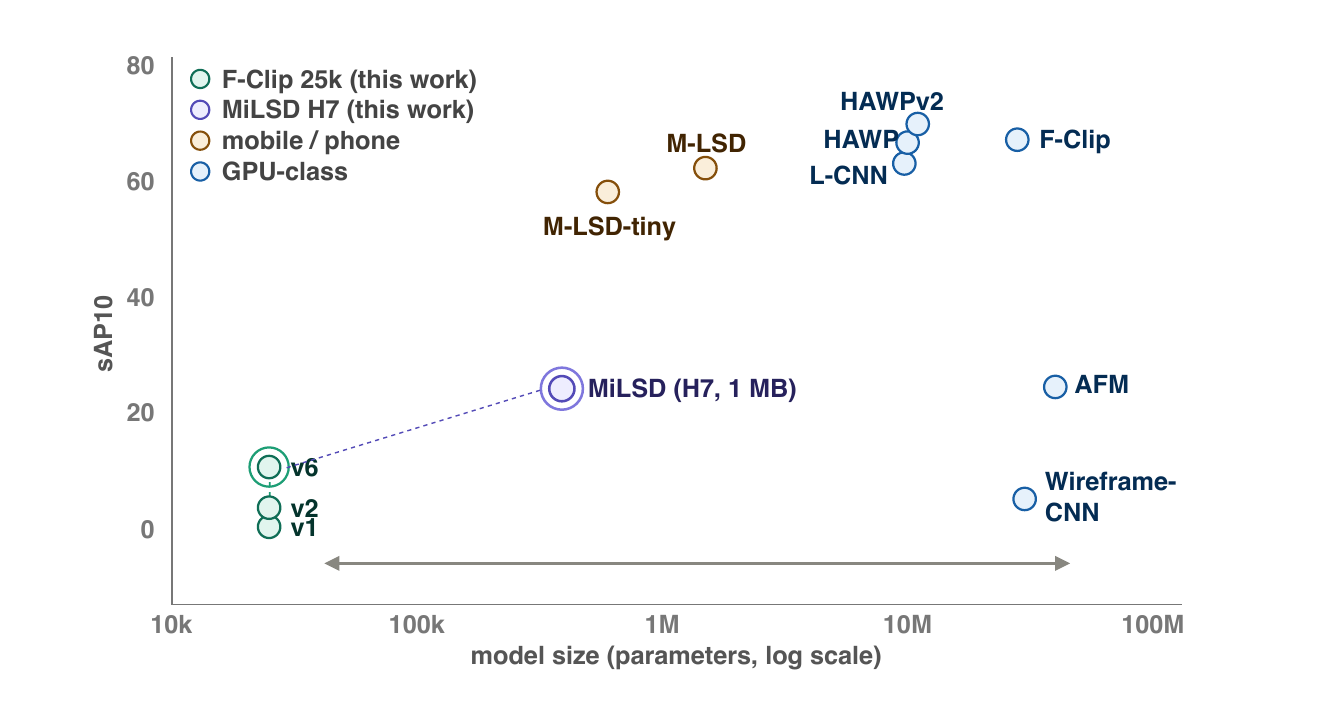}
\caption{Accuracy vs.\ parameter budget on the Wireframe benchmark (log-scale $x$).
Our two operating points sit at the extreme low-resource end: the 25k-parameter
F-Clip model on the STM32F746 ($\sAP{10}=10.6$) and MiLSD on the STM32H7 ($0.39$\,M
parameters, $\sAP{10}=24.1$, purple). Related learned detectors use $24\times$ to
$1{,}600\times$ more parameters and assume mobile or GPU compute; their figures are
from the respective papers (Table~\ref{tab:related})~\cite{xue2023hawpv2,dai2021fclip,xue2020hawp,zhou2019lcnn,gu2022mlsd,xue2019afm,huang2018wireframe}.}
\label{fig:accuracy_memory}
\end{figure}

Table~\ref{tab:related} and Fig.~\ref{fig:accuracy_memory} jointly situate our
two operating points within the broader accuracy--resource frontier of the
learned line-segment detection literature. At the low-resource end of
this range, the 25k-parameter F746 model achieves $\sAP{10}=10.6$ within a
0.25\,MB activation footprint, while MiLSD (Section~\ref{sec:milsd}) extends
this capability to $\sAP{10}=24.1$ under the expanded 1\,MB SRAM budget of the
STM32H7. In absolute accuracy terms, both models remain substantially below the
performance of contemporary transformer-era parsers, including DT-LSD at
$71.7$~\cite{dtlsd} and LINEA at $65.0$--$67.9$~\cite{linea}, as well as
compact GPU-oriented designs such as EM-LSD, which attains $63.2$ with
$1.1$\,M parameters~\cite{em-lsd}. This disparity is an expected consequence of
operating in a memory regime where peak SRAM is measured in kilobytes rather
than megabytes. On the resource axis, however, our position is distinctive:
Table~\ref{tab:related} lists no other learned detector with an affirmative
\emph{On MCU?} entry or an accompanying int8/int4 quantization study. To our
knowledge, this is the first learned line-segment detector designed and
evaluated under sub-megabyte MCU SRAM budgets, occupying a previously empty
region between high-accuracy GPU-based parsers and classical lightweight
detectors.

\begin{table}[!t]
\caption{Accuracy vs.\ resources on Wireframe. Prior \sAP{10} figures are from the
respective papers or, for methods predating the sAP metric, from later
re-evaluations~\cite{dtlsd,xue2023hawpv2,linea,dai2021fclip,xue2020hawp,em-lsd,zhou2019lcnn,gu2022mlsd,xue2019afm,huang2018wireframe,vongioi2010lsd};
parameter counts are largely as tabulated by LINEA~\cite{linea}. LINEA reports the
462-image Wireframe validation split.}
\label{tab:related}
\centering
\small
\begin{tabular}{|p{3cm}|p{1cm}|p{1.5cm}|p{0.8cm}|}
\toprule
Method & \sAP{10} & Params & On MCU?\\
\midrule
DT-LSD~\cite{dtlsd}  & 71.7 & 217\,M      & no\\
HAWPv2~\cite{xue2023hawpv2}     & 69.7 & $\sim$11\,M  & no\\
LINEA-L~\cite{linea} & 67.9 & 25\,M       & no\\
F-Clip~\cite{dai2021fclip}      & 67.4 & $\sim$28\,M  & no\\
HAWP~\cite{xue2020hawp}         & 66.5 & $\sim$10\,M  & no\\
LINEA-N~\cite{linea} & 65.0 & 3.9\,M      & no\\
EM-LSD~\cite{em-lsd}       & 63.2 & 1.1\,M      & no\\
L-CNN~\cite{zhou2019lcnn}       & 62.9 & $\sim$9.7\,M & no\\
M-LSD~\cite{gu2022mlsd}         & 62.1 & 1.5\,M       & no\\
M-LSD-tiny~\cite{gu2022mlsd}    & 58.0 & 0.6\,M ($\ge$78\,MB) & no\\
AFM~\cite{xue2019afm}           & 24.4 & $\sim$43\,M  & no\\
Wireframe-CNN~\cite{huang2018wireframe} & 5.1 & $\sim$30\,M & no\\
Classical LSD~\cite{vongioi2010lsd} & $\approx$0 & n/a & CPU\\
\midrule
\textbf{Ours: F-Clip int8 (F746)} & 10.6 & 0.025\,M (0.25\,MB) & yes\\
\textbf{Ours: MiLSD (H7)}          & \textbf{24.1} & \textbf{0.39\,M ($\sim$1\,MB)} & \textbf{yes}\\
\bottomrule
\end{tabular}
\end{table}

\subsection{Qualitative Results}\label{sec:res-qual}

\begin{figure*}[!t]
\centering
\includegraphics[width=0.92\textwidth]{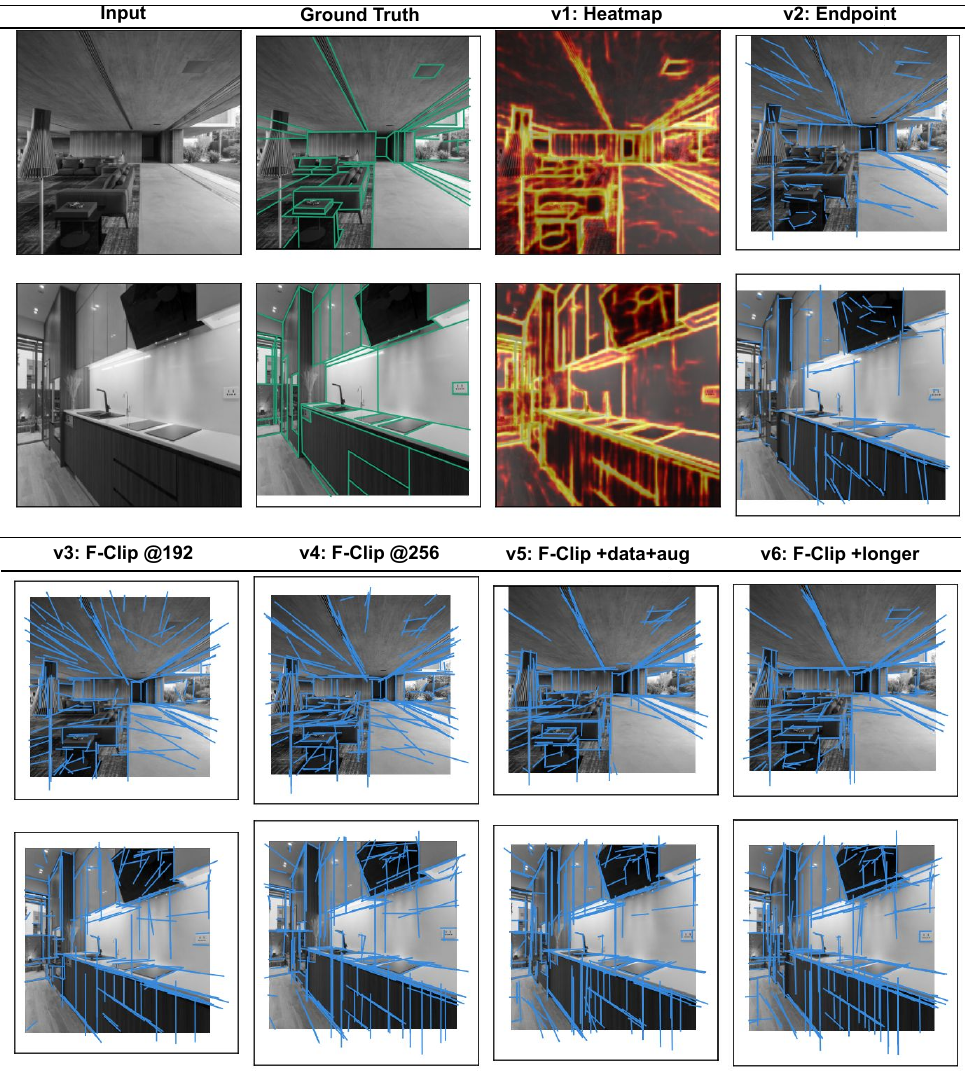}
\caption{Detections on two Wireframe-val images for v1 (heatmap), v2 (endpoint),
and v3--v6 (F-Clip progression); layout described in
Section~\ref{sec:res-qual}. The dominant visual step is v2$\rightarrow$v3; later
versions refine segment placement incrementally.}
\label{fig:versions}
\end{figure*}

Fig.~\ref{fig:versions} provides a qualitative side-by-side comparison of
versions v1 through v6 on two representative Wireframe-val
images~\cite{lsdtmlkaggle}, offering visual corroboration of the quantitative
trends reported above. In the heatmap formulation (v1), predictions remain
densely distributed across the image without resolving into discrete, well-formed
segments. Endpoint regression (v2) produces scattered short segments that fail to
reconstruct the underlying room structure. Beginning with F-Clip at v3, the
detections progressively recover coherent architectural geometry, a visual
pattern that directly mirrors the quantitative $\sAP{10}$ jump at v3
($3.6\rightarrow7.2$) and the more incremental refinement observed across
v3--v6 ($7.2\rightarrow10.6$). The qualitative contrast is visually striking:
F-Clip yields coherent, well-localized segments that align closely with salient
architectural edges, whereas the heatmap and endpoint alternatives continue to
produce noisy, fragmented, or incomplete detections that lack geometric
consistency.

\section{MiLSD: Verification-Augmented Detection on the STM32H7}\label{sec:milsd}

\begin{figure*}[!t]
\centering
\includegraphics[trim={0.5cm 0cm 0.5cm 0.1cm},clip,width=\textwidth,keepaspectratio]{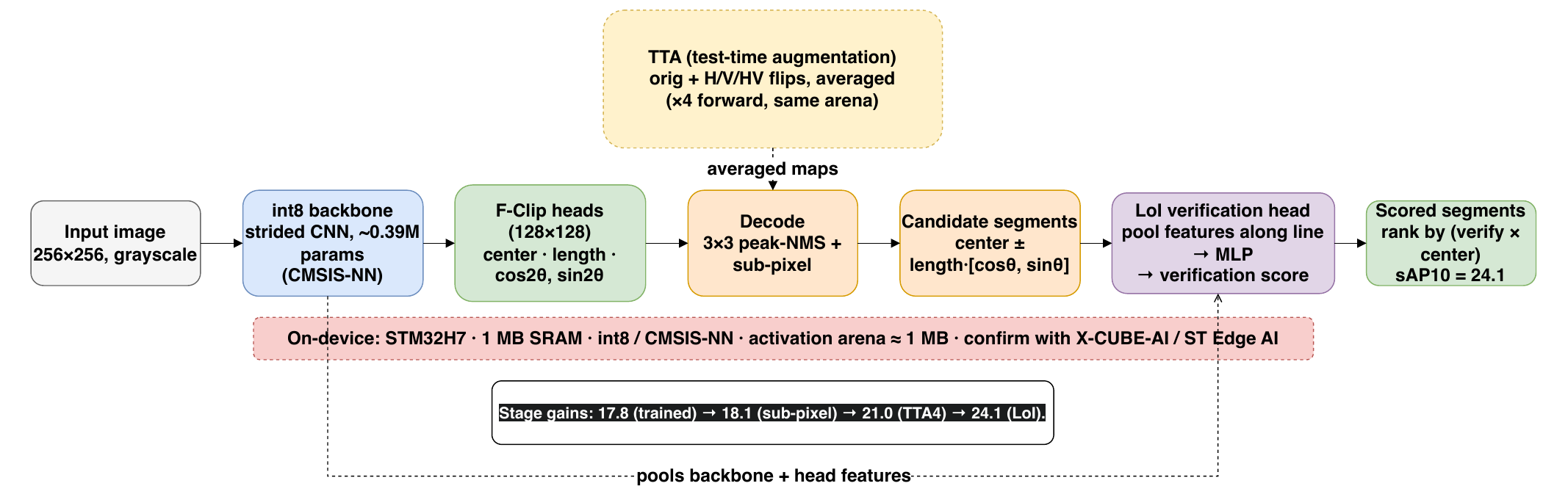}
\caption{MiLSD inference pipeline on the STM32H7. An int8 fully-convolutional
backbone predicts F-Clip center/length/angle maps; candidate segments are decoded
by $3{\times}3$ peak non-maximum suppression with sub-pixel refinement; optional
test-time flip augmentation averages the predicted maps; and a small
Line-of-Interest (LoI) head pools features along each candidate and re-scores it.
All stages reuse a single ${\approx}1$\,MB activation arena.}
\label{fig:milsd-pipe}
\end{figure*}

The 320\,KB SRAM budget of the STM32F746 imposes a hard ceiling on the detector
developed in Section~\ref{sec:method}, restricting it to approximately 25k
parameters and a correspondingly minimal activation footprint. A larger yet still
microcontroller-class platform, the STM32H7, which provides 1\,MB of SRAM and is
executed via CMSIS-NN~\cite{lai2018cmsisnn} on a Cortex-M7 core, affords
sufficient headroom to accommodate both a more capable backbone and a
substantially richer inference pipeline. We designate the resulting system
\emph{MiLSD} (Micro Line-Segment Detector). MiLSD preserves the F-Clip
center--length--angle output representation~\cite{dai2021fclip}
(Section~\ref{sec:repr}) introduced earlier and extends it along two
complementary axes: first, a capacity-scaled int8 backbone whose activation arena
is deliberately sized to approach, but not exceed, the 1\,MB SRAM budget; and
second, a sequence of inference-time refinement stages, comprising sub-pixel
decoding, test-time augmentation, and a learned verification head, each of
which contributes additional accuracy without increasing the size of the trained
network. The complete pipeline is illustrated in
Fig.~\ref{fig:milsd-pipe}. In contrast to the F746 model, whose performance is
bounded by its severely constrained 25k-parameter budget, MiLSD exploits the
expanded memory envelope to scale model capacity while preserving a lightweight,
inference-efficient pipeline appropriate for real-time embedded deployment.

\subsection{Capacity-Scaled Backbone}

With eight times the SRAM of the F746, the backbone width is increased until the
int8 activation arena approaches, but does not exceed, 1\,MB, yielding a
approximately 0.39\,M-parameter model (peak arena approximately 1\,MB at 256\,px input; a
0.7\,MB fallback configuration is also retained for devices with tighter memory).
Trained as in Section~\ref{sec:method} for 300 epochs, this model attains
$\sAP{10}=17.8$ on Wireframe-val, versus $10.6$ for the 25k F746 model,
confirming that, beyond the extreme F746 regime, capacity is a genuine lever.
This 68\% relative improvement demonstrates that the additional parameters are
effectively utilized when the memory budget permits, validating our decision to
scale the backbone for the H7 platform.

\subsection{Sub-Pixel Decoding and Test-Time Augmentation}

The center peaks are refined to sub-pixel accuracy by a one-dimensional parabolic
fit over each peak's neighbourhood, directly targeting the endpoint precision that
structural AP rewards ($\sAP{10}: 17.8 \rightarrow 18.1$). This refinement is
particularly beneficial for structural AP, which penalizes endpoint localization
errors quadratically. Averaging the predicted maps over the image and its
horizontal, vertical, and diagonal flips (test-time augmentation, TTA) further
improves robustness ($\sAP{10}: 18.1 \rightarrow 21.0$), a gain of nearly 3 points
from the four-view ensemble. TTA multiplies inference latency by the number of
views but reuses the same activation arena, so it does not raise peak memory,
making it a memory-free accuracy boost at the cost of increased latency.

\subsection{Line-of-Interest Verification Head}

The largest gain comes from a trained verifier in the spirit of
L-CNN~\cite{zhou2019lcnn} and HAWP~\cite{xue2020hawp}, scaled to the
microcontroller. With the backbone \emph{frozen}, a small multilayer perceptron
takes, for each candidate segment, features bilinearly pooled at 32 points along
the line from the backbone feature map and the output maps, together with a short
geometric descriptor, and predicts a verification score (real vs.\ spurious),
trained with one-to-one matched labels. Re-ranking by verification $\times$ center
score reaches the MiLSD row of Table~\ref{tab:milsd}. The decode and post-filtering
design were obtained by analyzing the inference stages of the principal wireframe
parsers~\cite{zhou2019lcnn,xue2020hawp,dai2021fclip,zhang2019ppgnet,zhao2020dht};
across that study, weight-free post-processing saturates near $\sAP{10}=21$, and
only the trained verifier advances beyond it, contributing an additional 3.1 points
to reach 24.1.

\begin{table}[tb]
\centering
\caption{MiLSD on Wireframe-val (STM32H7 model). Each stage is inference-time only;
the trained network is unchanged. The oracle ranks the same candidates by their
true labels and is the recall ceiling of the candidate set.}
\label{tab:milsd}
\begin{tabular}{lccc}
\toprule
Configuration & \sAP{5} & \sAP{10} & \sAP{15}\\
\midrule
Trained model (sub-pixel decode)          & 12.1 & 18.1 & 21.2\\
\quad + test-time augmentation ($\times4$) & 15.1 & 21.0 & 24.0\\
\quad + LoI verification (\textbf{MiLSD})  & \textbf{16.0} & \textbf{24.1} & \textbf{27.9}\\
\midrule
\emph{Oracle (perfect verifier)}          & 24.1 & 37.4 & 40.0\\
\bottomrule
\end{tabular}
\end{table}

The oracle row ($\sAP{10}=37.4$) bounds what the candidate set can deliver;
the LoI head recovers about 55\% of the gap from TTA. The remainder reflects
unannotated image edges and recall limits; junction-based candidate
generation~\cite{zhou2019lcnn} could help but exceeds the 1\,MB budget.

\subsection{Qualitative Results}

\begin{figure}[!t]
\centering
\includegraphics[width=\columnwidth,keepaspectratio]{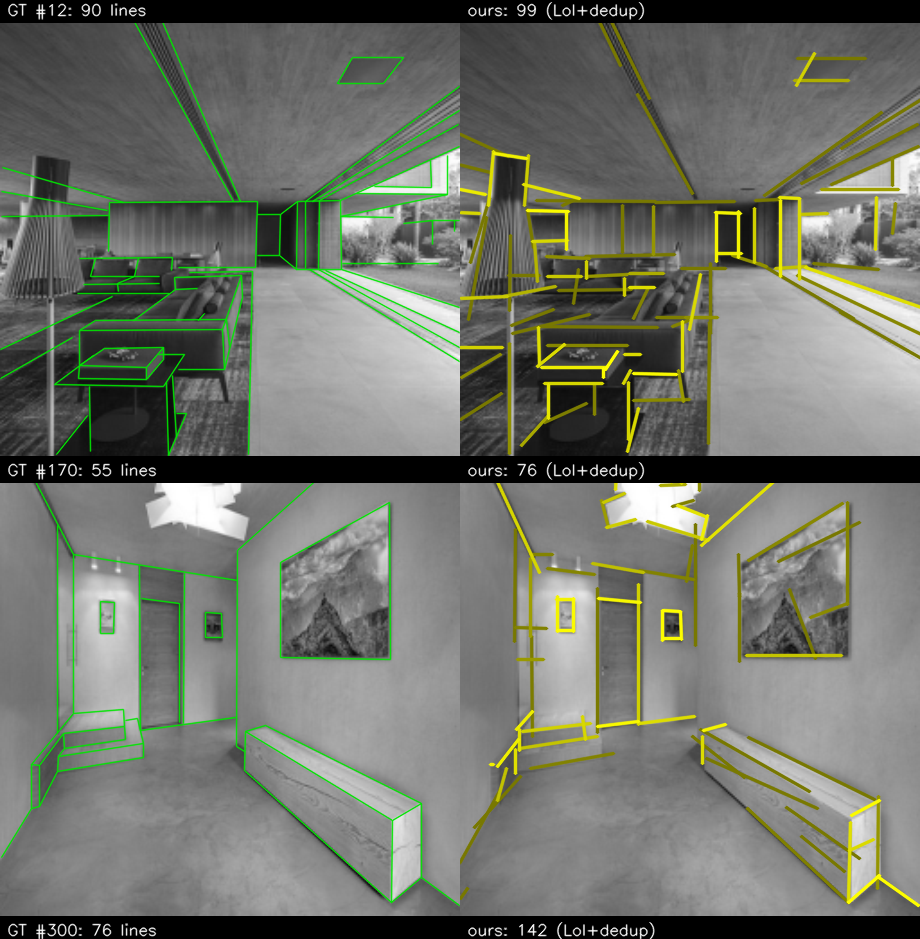}
\caption{Four Wireframe-val scenes: ground truth (left, green) and MiLSD output
(right, yellow) after LoI verification; line opacity encodes verifier confidence.
}
\label{fig:milsd-ex}
\vspace{-1em}
\end{figure}

Fig.~\ref{fig:milsd-ex} shows MiLSD on held-out scenes. The detector is complete on
the salient structure (cabinetry, counters, window mullions, and architectural
edges are recovered at correct orientation and length). The dense scene of example~2
illustrates the verifier suppressing the redundant ridge detections that a raw
center-heatmap emits, demonstrating the effectiveness of the learned verification
head. Example~3 shows that the remaining stray predictions are real but unannotated
edges (reflections, texture seams), consistent with the oracle analysis above,
which identified that many false positives are actually unlabeled ground-truth
edges. These qualitative results validate that MiLSD achieves its accuracy gains
through meaningful structural understanding rather than overfitting to the
training set.

\subsection{Deployment Budget}

The SRAM ceiling is set by the backbone's int8 activation arena (approximately 1\,MB);
the LoI head's pooled feature ($<$0.5\,MB) fits within that peak and TTA reuses it,
so neither raises it. Weights total approximately 0.54\,MB (backbone plus head), well
within flash. Latency is dominated by the int8 convolutions and scales with the
number of TTA views; the LoI head adds a small per-line cost. On-device arena and
latency must be confirmed with ST Edge AI~\cite{stedgeai}; the figures here are
software-measured. The memory-efficient design ensures that all inference stages
operate within the 1\,MB SRAM budget, making MiLSD deployable on the STM32H7
without requiring external memory. Source code is released
for reproducibility \cite{milsd2026code}.
\section{Conclusion}\label{sec:conclusion}

This paper presented MiLSD, a line-segment detector explicitly designed for microcontroller-scale memory, alongside a systematic study of the accuracy--resource trade-off under extreme memory constraints. We compared three output representations and found that the F-Clip center-with-length-and-angle encoding learns most effectively at small model sizes, achieving $\sAP{10}=10.6$ with only 25k parameters. Our quantization study revealed that 8-bit weights preserve full-precision accuracy, while 4-bit quantization collapses, particularly in the $(\cos2\theta,\sin2\theta)$ angle regression, with quantization-aware training recovering only part of the loss. By scaling the backbone to a 1\,MB activation budget and adding inference-time enhancements including sub-pixel decoding, test-time augmentation, and a Line-of-Interest verification head, MiLSD improves $\sAP{10}$ from $10.6$ at 0.25\,MB to $24.1$ within 1\,MB on the ShanghaiTech Wireframe benchmark.

To our knowledge, no prior work characterizes the joint trade-off among representation choice, quantization bit-width, and on-device post-processing at this memory scale. F-Clip wins at int8 but its angle head is int4-fragile, whereas heatmaps tolerate lower precision yet require an external linker, a finding that informs future TinyML geometric-vision designs. As expected, Wireframe accuracy remains far below GPU-class parsers given the SRAM envelope we target; that gap is inherent to the memory regime rather than a limitation we aim to overcome.


\FloatBarrier
\bibliographystyle{IEEEtran}
\bibliography{References}

\begin{IEEEbiography}[{\includegraphics[width=1in,height=1.25in,clip,keepaspectratio]{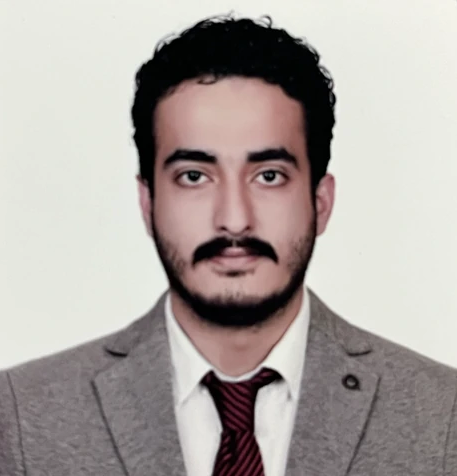}}]{Parsa Hassani Shariat Panahi}
received his B.Sc. degree in Computer Engineering from Azad University South Tehran Branch, Tehran, Iran, and the M.Sc. degree in Computer Engineering - Computer Networks from Iran University of Science and Technology, Tehran, Iran. His research interests include cellular networks, QoE assessment, telecommunication networks, wireless communication, and machine learning. He can be reached at parsa\_hassani@comp.iust.ac.ir.
\end{IEEEbiography}
\vspace{-4em}
\begin{IEEEbiography}[{\includegraphics[width=1in,height=1.25in,clip,keepaspectratio]{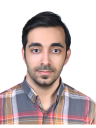}}]{Amir Hossein Jalilvand}
received his B.Sc. degree in Computer Engineering from Bu-Ali Sina University, Hamadan, Iran, and the M.Sc. degree in Computer Engineering - Computer Architecture from Iran University of Science and Technology, Tehran, Iran. He is currently pursuing his Ph.D. in Computer Engineering. His research interests include cellular networks, stochastic and unary computing, computer architecture, fuzzy logic, and machine learning. Mr. Jalilvand has authored several publications in these fields. He can be reached at jalilvand\_a@comp.iust.ac.ir.
\end{IEEEbiography}
\vspace{-4em}
\begin{IEEEbiography}[{\includegraphics[width=1in,height=1.25in,clip,keepaspectratio]{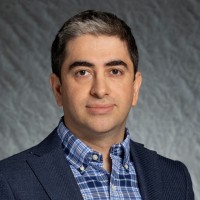}}]{M. Hassan Najafi}
received his Ph.D. in electrical and electronics engineering from the University of Minnesota-Twin Cities, Minneapolis, MN, USA, in 2018. He is currently an Associate Professor at the Electrical, Computer, and Systems Engineering Department at Case Western Reserve University. His research interests include stochastic and approximate computing, unary processing, in-memory computing, and hyperdimensional computing. He has authored/coauthored more than 120 peer-reviewed papers and has been granted 12 U.S. patents with more pending. Dr. Najafi received the NSF CAREER Award in 2024, the Best Paper Award at GLSVLSI'23 and ICCD'17, and the 2018 EDAA Outstanding Dissertation Award. Dr. Najafi is a senior member of IEEE and a senior member of the U.S. National Academy of Inventors (NAI). He can be reached at najafi@case.edu.
\end{IEEEbiography}

\end{document}